%% file: root.tex
\documentclass[letterpaper, 10 pt, conference]{ieeeconf}  

\IEEEoverridecommandlockouts                              

\usepackage{graphicx}
\usepackage{algorithmic}
\usepackage{algorithm}
\usepackage{amsmath}
\usepackage{amssymb}
\usepackage{multirow}
\usepackage[table,xcdraw]{xcolor}
\usepackage{hyperref}

\title{\LARGE \bf
ManiAgent: An Agentic Framework for General Robotic Manipulation
}

\author{Yi Yang$^{1, \dag}$, Kefan Gu$^{2, \dag}$, Yuqing Wen$^{3, \dag}$, Hebei Li$^{3, \dag}$, Yucheng Zhao$^{4, \ddag}$, Tiancai Wang$^{4}$ and Xudong Liu$^{1, *}$
\thanks{$^{\dag}$: This work was done during the internship at Dexmal.}
\thanks{$^{\ddag}$: Project lead.}%
\thanks{$^{*}$: Corresponding author.}%
\thanks{$^{1}$ Yi Yang and Xudong Liu are with Beijing University of Technology.}%
\thanks{$^{2}$ Kefan Gu is with Nanjing University.}%
\thanks{$^{3}$ Yuqing Wen and Hebei Li are with University of Science and Technology of China.}%
\thanks{$^{4}$ Yucheng Zhao and Tiancai Wang are with Dexmal.}%
}

\begin{document}

\maketitle
\thispagestyle{empty}
\pagestyle{empty}


\input{sections/0_abs}
\input{sections/1_intro}
\input{sections/2_related}
\input{sections/3_method}
\input{sections/4_exp}
\input{sections/5_con}

\addtolength{\textheight}{-12cm}   

\input{sections/6_sup}
\end{document}

%% file: sections/0_abs.tex
\begin{abstract}

While Vision-Language-Action (VLA) models have demonstrated impressive capabilities in robotic manipulation, their performance in complex reasoning and long-horizon task planning is limited by data scarcity and model capacity. To address this, we introduce ManiAgent, an agentic architecture for general manipulation tasks that achieves end-to-end output from task descriptions and environmental inputs to robotic manipulation actions. In this framework, multiple agents involve inter-agent communication to perform environmental perception, sub-task decomposition and action generation, enabling efficient handling of complex manipulation scenarios. Evaluations show ManiAgent achieves an 86.8\% success rate on the SimplerEnv benchmark and 95.8\% on real-world pick-and-place tasks, enabling efficient data collection that yields VLA models with performance comparable to those trained on human-annotated datasets. The project webpage is available at \href{https://yi-yang929.github.io/ManiAgent/}{https://yi-yang929.github.io/ManiAgent/}.

\end{abstract}

%% file: sections/1_intro.tex
\section{Introduction}
Recent advances in large language models (LLMs) have brought capabilities to the field of robot manipulation. Among these, Vision-Language-Action (VLA) models have emerged as a promising and active approach~\cite{li2024cogact,black2024pi_0,brohan2023rt,kim2024openvla}, leveraging LLMs for end-to-end robotic control. By fine-tuning LLMs on robotic demonstration data through imitation learning, VLAs integrate vision, language, and action modalities, enabling robots to perform diverse tasks from natural language instructions and visual observations~\cite{belkhale2024rth}. Despite these advances, current VLAs suffer from two critical limitations: a strong dependence on large-scale, high-quality robotic data and insufficient task intelligence when facing complex scenarios. In terms of data, VLAs heavily rely on high-quality demonstrations, which are costly to collect and still insufficient to cover the diversity of real-world environments, leading to significant performance degradation under data-scarce or out-of-distribution(OOD) conditions~\cite{black2024pi_0,kim2024openvla}. Moreover, their task intelligence remains limited when handling complex tasks. For example, fine-tuning on robotic data often erodes the LLM’s original high-level understanding, making it difficult for the model to interpret indirect instructions or perform complex reasoning. Similarly, for long-horizon tasks, VLAs exhibit limitations in the advanced planning required to successfully complete these tasks.

\begin{figure}[htbp]
    \centering
    \includegraphics[width=1\linewidth]{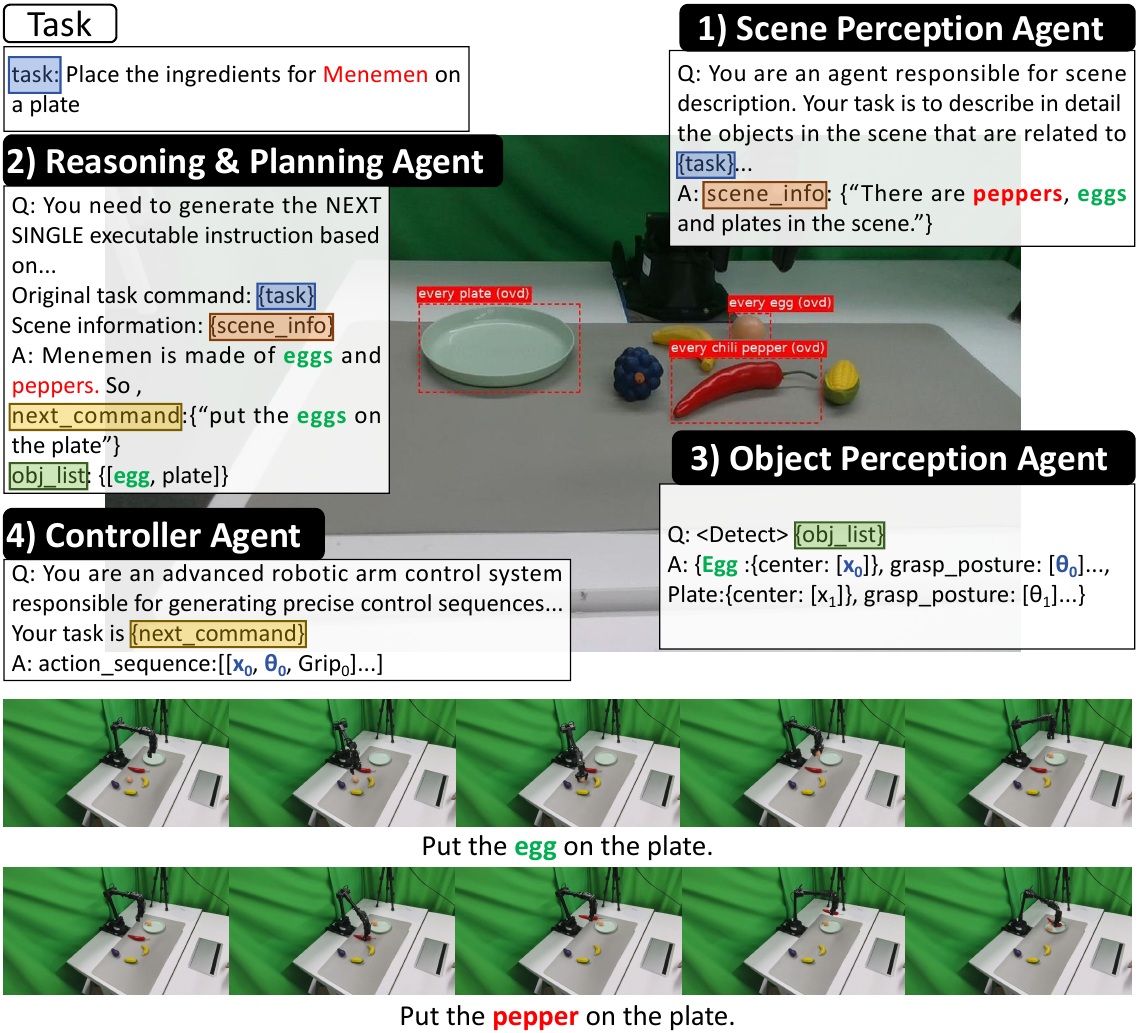}
    \vspace{-2em}
    \caption{\textbf{ManiAgent} decomposes the Menemen(a pepper-and-egg dish) ingredient-finding task into perception, reasoning, and execution handled by dedicated agents.}
    \label{fig:figure1}
    \vspace{-2em}
\end{figure}
Motivated by the limitations of current VLAs, in this paper, we propose a training-free, end-to-end framework for robot manipulation that leverages powerful LLMs in an agentic manner. To empower the framework to accomplish complex tasks, we introduce an Agent-based general robotic Manipulation framework (ManiAgent), which decomposes tasks and distributes them to specialized agents for effective problem solving. ManiAgent consists of three specialized agents that form a complete perception–reasoning–control pipeline. Specifically, a perception agent extracts detailed spatial information from the physical environment, including scenes and objects; a reasoning agent performs intention reasoning and sub-task decomposition, providing high-level planning for long-horizon tasks; and an action-execution agent integrates perceptual and instruction information to generate robot actions directly executable in the real world. Through the seamless collaboration of these agents, ManiAgent can accomplish a wide range of complex manipulation tasks with high success rates, without requiring any training data. Furthermore, to enhance the efficiency of our framework, we design a dedicated caching mechanism that leverages parameterized action sequences as caches, thereby accelerating task completion.

Extensive experiments conducted in both simulation and real-world environments demonstrate the superior performance of ManiAgent. In SimplerEnv~\cite{li2024evaluating}, ManiAgent achieves a success rate of 86.8\%, representing a substantial improvement over CogACT’s 51.3\%~\cite{li2024cogact}. In real-world experiments, our method, when combined with a VLM (Vision-Language Model) of strong capability, achieves a 100\% success rate on complex multi-step reasoning tasks (e.g., placing the knife and fork in the right hands based on table etiquette.). Given ManiAgent’s high success rate in physical scenarios, it can also serve as an effective data collection tool, enabling the automated generation of large-scale training data with minimal human intervention. It is demonstrated that VLA models trained on data collected by ManiAgent achieve performance comparable to those trained on human-collected datasets, significantly reducing labor costs and providing strong data support for learning-based approaches such as VLAs.

In summary, our key contributions are as follows:
\begin{itemize}
    \item We propose an end-to-end agentic framework named ManiAgent that directly generates executable action sequences for general robot manipulation tasks.
    \item We design a perception–reasoning–control pipeline by coordinating three specialized agents, integrating spatial perception, task reasoning, and action planning to enable complex task execution.
    \item We conduct extensive experiments to validate ManiAgent’s superior performance. Its high success rate allows it to serve as a fully automated data collection tool, providing strong support for other learning-based robot manipulation methods.
\end{itemize}

%% file: sections/2_related.tex
\section{Related Work}
\label{sec:related}
\subsection{Vision-Language-Action Models}

Vision-Language-Action (VLA) models represent a significant advancement in robotic manipulation by building on large language models (LLMs) and integrating multimodal inputs for direct action prediction. Pioneering works like Unleashing VLA~\cite{wu2023unleashing} and RT-2~\cite{brohan2023rt2} demonstrate how these models map vision and language to actions, enabling end-to-end deployment in controlled settings. CogACT~\cite{li2024cogact} emphasizes the continuous, temporal, and precise characteristics of actions, separating cognition from execution and focusing on action sequence modeling. Similarly, Pi0~\cite{black2024pi_0} explores zero-shot policy adaptation for manipulation, leveraging pretrained models for out-of-distribution tasks. Other notable contributions include ROSA~\cite{wen2025rosa} enables the VLA model to gain enhanced spatial understanding and self-awareness, Exploring VLA~\cite{zhou2025exploring} for hierarchical task analysis, RoboCerebra~\cite{han2025robocerebra} for brain-inspired action planning, and Towards VLA~\cite{li2024towards} addressing temporal bottlenecks. 

Although VLAs have achieved significant progress, they remain heavily dependent on large-scale data and lack sufficient task intelligence to handle complex scenarios, such as intention reasoning, long-horizon tasks, and advanced planning. In contrast, our work takes a different approach by leveraging the inherent capabilities of LLMs in an agentic manner, resulting in a training-free, end-to-end framework that can accomplish diverse and complex tasks without requiring additional demonstration data.

\subsection{Agent-based Frameworks for Robot Manipulation}

Agent-based frameworks extend LLMs to environmental interactions, often starting with tool-augmented agents like ToolLLM~\cite{qin2024toolllm} for API mastery, StableToolBench~\cite{guo2024stabletoolbench} for benchmarking, ToLeaP~\cite{chen2025toleap} for optimized integration, and EASYTOOL~\cite{yuan2024easytool} for simplified instructions and performance gains.

Building on these foundations, single-agent methods adapt tools for general manipulation tasks. For instance, compositional approaches like VoxPoser~\cite{huang2023voxposer} generate trajectories via 3D value maps, while constraint-based methods such as ReKep~\cite{huang2024rekep} use keypoints and GeoManip~\cite{tang2025geomanip} employ geometric interfaces. Code-generation techniques, including ProgPrompt~\cite{singh2023progprompt}, Code as Policies~\cite{liang2023code}, and CodeDiffuser~\cite{yin2025codediffuser}, produce executable policies to handle sequencing and ambiguity. Other works like Manipulate-Anything~\cite{duan2024manipulateanything} enable automation through waypoints, and SayPlan~\cite{rana2023sayplan} grounds planning with scene graphs. However, these approaches often lack robust supervision, resulting in limited closed-loop feedback for complex, dynamic tasks.

To improve task supervision and refinement, frameworks incorporating agent interactions enable collaborative intelligence and adaptive dynamics. Examples include Internet of Agents~\cite{chen2025internet} for connecting heterogeneous capabilities and Evolving Agents~\cite{li2024evolving} for simulating personality-based evolution. In manipulation domains, systems like ReplanVLM~\cite{mei2024replanvlm}, ExploreVLM~\cite{lou2025explorevlm}, and MALMM~\cite{singh2024malmm} facilitate mutual oversight among components, achieving task-level closed loops through coordinated planning and execution.

Nevertheless, existing methods suffer from key limitations. Single-agent approaches often lack robust supervision and closed-loop feedback, limiting their performance on complex tasks. Interactive frameworks, facilitate coordination and adaptive reasoning, typically rely on manually defined, task-specific APIs, which restricts generalization and end-to-end deployment. In contrast, our ManiAgent leverages specialized agents for perception, planning, and execution to enable mutual oversight and adaptive refinement to directly generate robotic actions. This design eliminates the need for task-specific APIs, offering greater flexibility, easier deployment, and stronger generalization across diverse and complex manipulation scenarios, while also improving data efficiency and physical execution reliability compared to existing agent-based approaches.

%% file: sections/3_method.tex
\section{Framework of ManiAgent}
\label{sec:ManiAgent}
\begin{figure*}[htbp]  
    \centering
    \includegraphics[width=0.90\textwidth]{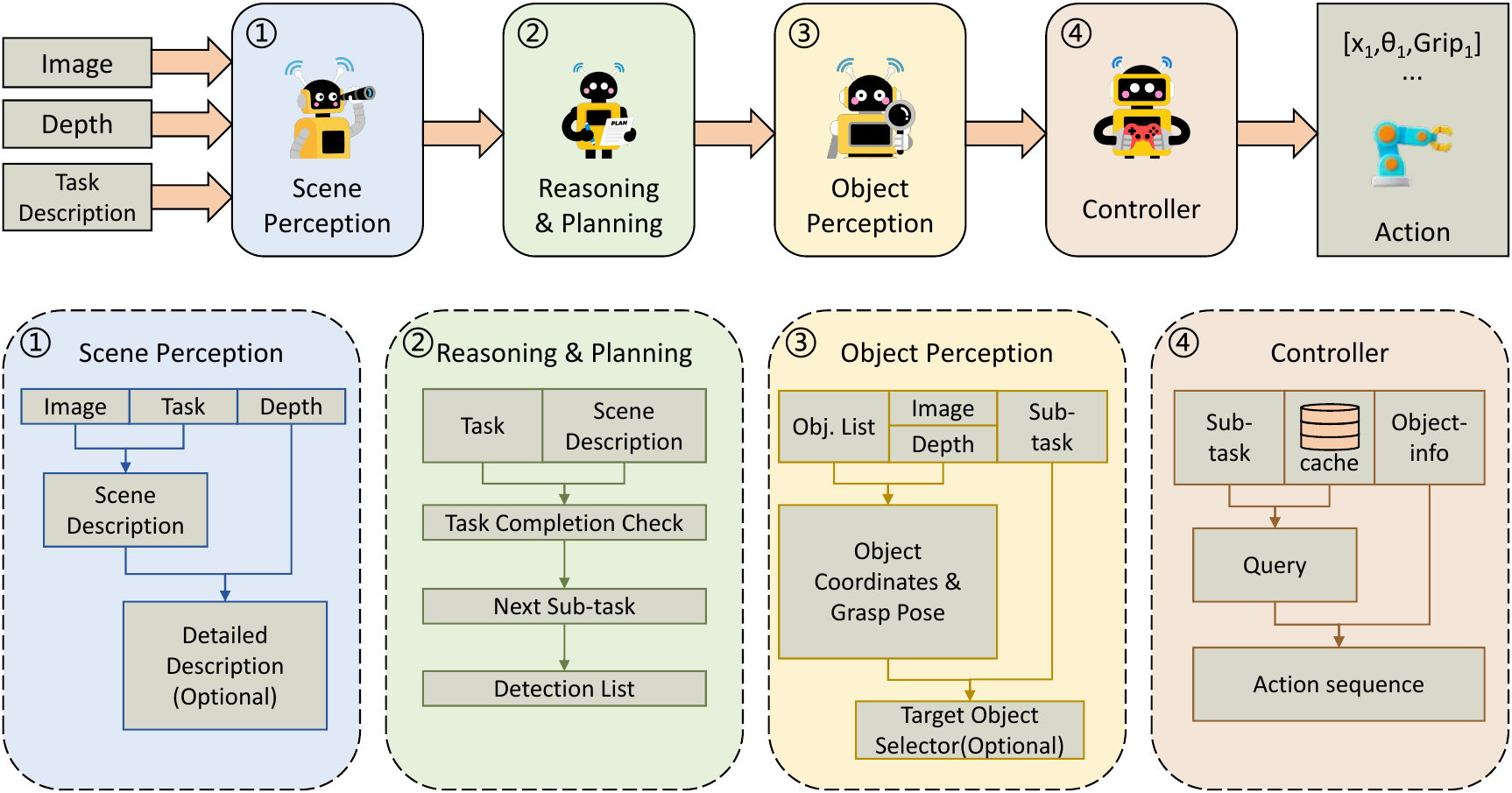}  
    \vspace{-0.5em}
    \caption{Overview of the \textbf{ManiAgent} framework. 1) The process begins with the perception agent, which takes scene images and user-provided instructions as input, and invokes a Vision-Language Model (VLM) to generate task-relevant scene descriptions. 2) The reasoning agent receives the scene descriptions and task instructions, then queries a Large Language Model (LLM) for status evaluation. 3) During sub-task execution, the perception agent uses object detection methods to identify target objects and retrieve detailed information. 4) The controller agent queries the cache based on the sub-task. If a matching cached action sequence is found, it is directly invoked; otherwise, the agent queries the LLM with the sub-task description and object details to generate a complete action sequence for execution.}
    \label{fig:workflow}  
    \vspace{-1.0em}  
\end{figure*}
In this section, we provide a detailed description of our proposed ManiAgent framework. As illustrated in Fig.~\ref{fig:workflow}, ManiAgent employs multiple specialized agents collaborating seamlessly to achieve end-to-end robotic manipulation. The process begins with the perception agent, which processes scene images and user instructions via a Vision-Language Model (VLM) to generate task-relevant scene descriptions. The reasoning agent then fuses these descriptions with task instructions to query a Large Language Model (LLM) and perform sub-task decomposition. During sub-task execution, the perception agent obtains detailed object information, including center positions and grasp poses. Finally, the controller agent assembles the object details and sub-task descriptions via the LLM to directly generate the requisite action sequences for completion.

\subsection{Scene Perception}
\label{sec:Scene}

The scene perception module utilizes a VLM to generate textual descriptions of the scene. The inputs to the VLM include the scene's image $I$, the task description $T$, and the prompts $\Theta$. After processing through the VLM, the output is a textual scene description $S$. This can be expressed as:
\begin{equation}
S = VLM(I, T; \Theta).
\end{equation}
The process of tuning the prompts $\Theta$ can be viewed as solving an optimization problem that balances key objectives:
\begin{equation}
\max_\Theta \mathcal{F}\bigl( \mathrm{Recall}(S, S_{\mathrm{true}}), \mathrm{Relevance}(S, T) \bigr),
\end{equation}
where $\mathcal{F}$ is a balancing function that optimizes both recall and relevance. In tuning the prompts, we prioritize ensuring that the scene description captures all task-relevant true information in the scene $S_{\mathrm{true}}$, thus focusing first on the recall between the output $S$ and $S_{\mathrm{true}}$, represented by $\mathrm{Recall}(S, S_{\mathrm{true}})$. After achieving this, to reduce the interference of redundant information on subsequent modules, we continue fine-tuning the prompts to maximize the relevance between the output scene description $S$ and the task description $T$, denoted by $\mathrm{Relevance}(S, T)$, ultimately finalizing the prompt settings for this module.

Upon completing scene perception, if the textual scene description is insufficient for the reasoning agent to plan the sub-task, object detection methods are invoked. By extracting key object information from the scene description and combining it with image input, the detection network identifies the objects and, using camera calibration parameters, computes their 3D spatial coordinates. This enhances the reasoning agent’s ability to incorporate 3D information, providing more detailed context for sub-task planning. For example, in a task that requires determining whether the pepper is at coordinates [0.1, 0.2, 0.0], VLMs like GPT-5 lack precise localization capabilities and cannot output accurate object coordinates in the image; thus, we invoke the detection method like Florence-v2~\cite{xiao2024florence} to identify the pepper's pixel coordinates and transform them into 3D scene coordinates via calibration parameters, enabling accurate task completion assessment.

\subsection{Reasoning and Planning}
\label{sec:Reasoning}

The Reasoning and Planning module serves as the backbone of the entire task execution process. It receives textual scene descriptions from the Scene Perception module, combines them with the physical knowledge embedded within the LLM, and takes into account the current progress of the task. Through this process, the reasoning agent decomposes the overall goal into smaller, executable subtasks. These subtasks are then communicated to the Object Perception module for object detection and the Controller module for execution. Furthermore, the reasoning agent compiles a list of objects that are crucial for completing each subtask, aiding the perception module in gathering the required object details for the next steps.

Initially, the reasoning agent evaluates the current state: if the previous sub-task succeeded or the process is in its initial state, it proceeds to the next step. For the obtained sub-task, the agent further decomposes its content to extract keywords that can be directly used for open-vocabulary detection. For instance, the sub-task of placing the pepper into the plate is decomposed into the pepper and the plate to facilitate open-vocabulary detection.

Notably, our sub-task decomposition is not performed all at once but incrementally, adapting step-by-step to the evolving scene. This ensures that generated sub-tasks are feasible within the current environment. During each planning iteration, the reasoning agent stores historical sub-tasks as memory to prevent local loops.

\subsection{Object Perception}
\label{sec:Object}
\begin{figure}[htbp]
    \vspace{-0em}
    \centering
    \includegraphics[width=1\linewidth]{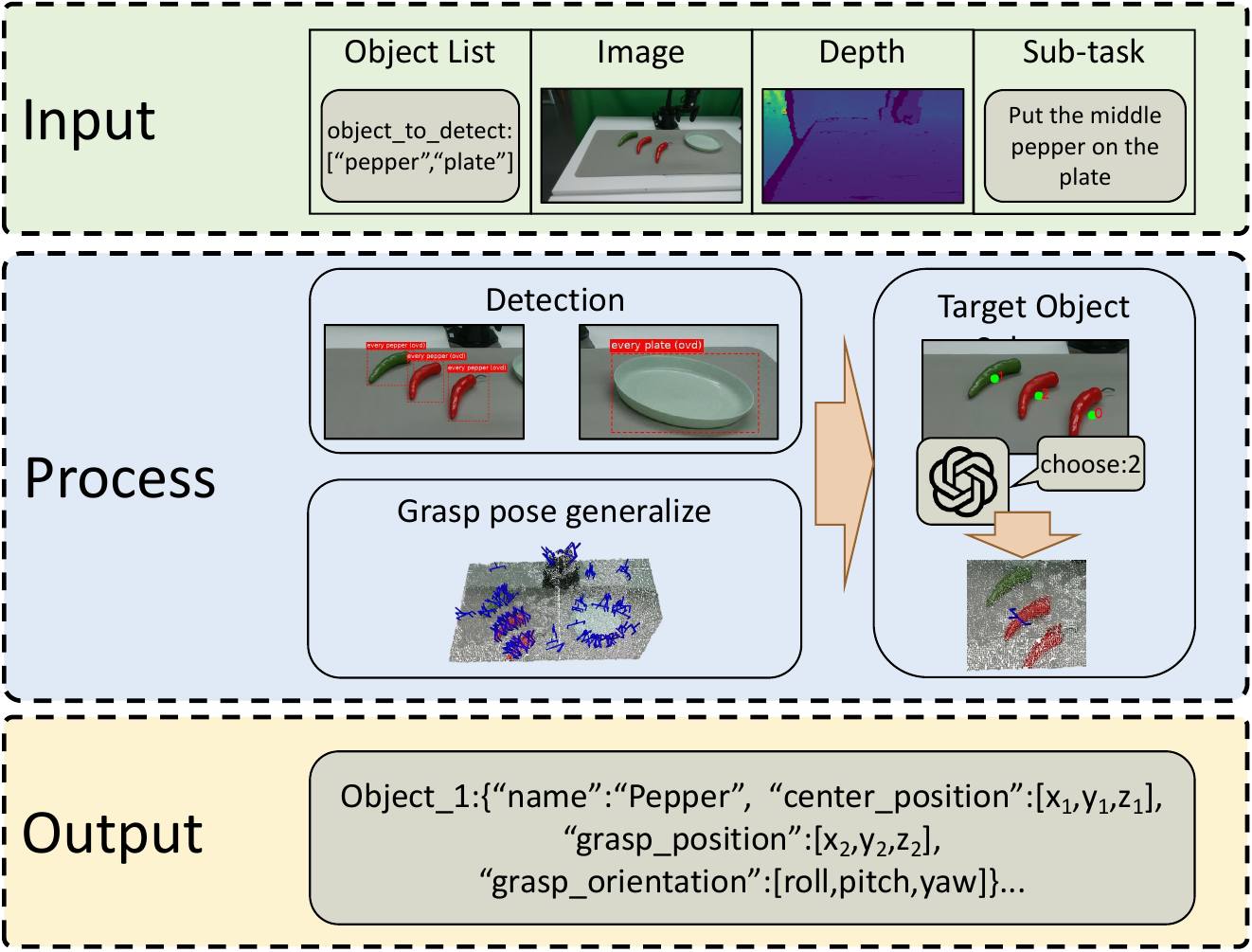}
    \vspace{-1.5em}
    \caption{The perception module of ManiAgent processes the target object list from the upper-level module with scene images, depth maps, and camera parameters to obtain object coordinates and grasping poses (using VLM for screening identical objects when needed), and finally sends text-format object information to the next module.}
    \label{fig:object_perception}
    \vspace{-0em}
\end{figure}

The perception module of ManiAgent first receives the target object list from the upper-level module, fuses the list with the scene image to obtain the target objects’ pixel coordinates, and then integrates the depth map and camera parameters to calculate their 3D coordinates. Meanwhile, a grasp pose generator produces grasping postures and positions for each object. For multiple identical objects in the scene, a Vision-Language Model (VLM) is invoked for further screening. Finally, the object information in text format is sent to the next module.

As shown in Fig.~\ref{fig:object_perception}, the object perception stage begins by utilizing a VLM with object detection capabilities~\cite{xiao2024florence} to perform open-vocabulary detection based on the pre-obtained list of objects to manipulate. When inputting the prompt, to ensure that all potential required items are detected, we added the word \textit{every} before the item being detected. This method effectively reduces the occurrence of missed detections. The VLM with object detection capabilities then provides the center points of the objects, which are determined as the centers of the bounding boxes. These image coordinates are subsequently transformed into 3D coordinates in the base coordinate system using intrinsics and extrinsics obtained from calibration between the camera and the robotic arm.

During this process, if multiple instances of the same object are detected, further confirmation of the target object is required. We annotate the center points of identical objects with sequential numbers on the image and input them to the VLM along with a tailored prompt to accurately identify the intended object, thereby accommodating specific object descriptions (e.g., the middle pepper).

Next, AnyGrasp~\cite{fang2023anygrasp} is employed to perceive grasp poses across the entire scene. By matching with the previously obtained object positions, the highest-scoring grasp position and pose within the neighborhood are selected as the grasping information for the object.

At the output stage, considering that both the grasp position and the center position of the object may be utilized by the following agents, we output both pieces of information.

\subsection{Controller}
\label{sec:Controller}
The controller agent is responsible for translating the sub-task received from the reasoning agent, along with corresponding scene images and depth information, into executable actions for the robotic arm. It directly outputs key points in Cartesian space and textual descriptions for each action step. This process ensures precise manipulation by generating or retrieving appropriate action sequences.

The action generation using the LLM can be formalized as solving a sequencing problem with known options, as expressed in Equation~\ref{eq:controller}:
\begin{equation}
A = \mathrm{LLM}\left(T, {(p_i, g_j) \mid p_i \in P, g_j \in G, \mathrm{Valid}(p_i, g_j, T)}\right),
\label{eq:controller}
\end{equation}
where \(A\) denotes the output set of action sequences. This set is constructed by the LLM through combining and ordering the center positions \(p_i\) and grasp poses \(g_j\) of specific objects based on the sub-task \(T\), with \(p_i \in P\) and \(g_j \in G\), where \(P\) and \(G\) are the sets of object center positions and grasp poses obtained from the object perception stage (Subsection~\ref{sec:Object}), respectively. During processing, the LLM ensures task relevance by selecting valid pairs via the \(\mathrm{Valid}(\cdot)\) function, filtering candidates that align with the sub-task \(T\). Our prompts are limited to a few basic skills (pick-and-place, dragging an object, and rotating an object), with one trajectory example per skill, which we believe maximizes the LLM's generalization performance. Throughout the input process, we rely exclusively on textual descriptions, as most scene information has already been deconstructed into precise object coordinates, eliminating the need for additional image inputs.

However, LLM-based action generation may introduce latency issues, to mitigate this, we employ a caching mechanism. For subtasks previously executed and cached, the agent first queries the cache to determine whether existing parameterized action sequences match the current task. If matching records exist, the agent retrieves the cached parameterized action sequence and integrates it with details of key objects in the current scene, such as coordinates, to produce specific actions. This integration relies on fixed object indices in the scene information, a consistency ensured by the object perception stage described in Subsection~\ref{sec:Object}. Upon completing a new task, the controller outputs a parameterized action sequence for caching, which is stored as a list and can be retrieved when a future task prompt exactly matches an existing cached prompt.

%% file: sections/4_exp.tex
\section{Experiment}
\subsection{Simulation Setup}

\begin{figure}[htbp]
    \vspace{-1em}
    \centering
    \includegraphics[width=0.7\linewidth]{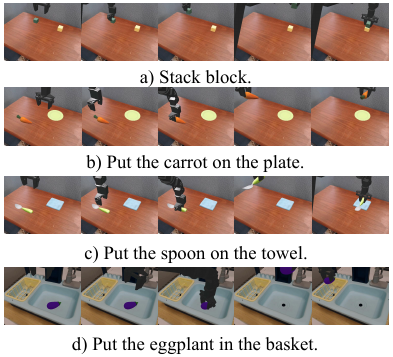}
    \vspace{-0.5em}
    \caption{Task execution process in the simplerenv simulation environment}
    \label{fig:simulation_image}
    \vspace{-0.5em}
\end{figure}

To systematically evaluate the capability of our method, we first conducted experiments on the SimplerEnv~\cite{li2024evaluating} platform. As illustrated in Fig.~\ref{fig:simulation_image}, we selected the widely recognized BridgeTable-v1 and BridgeTable-v2 as experimental environments, with the corresponding tasks listed below:
\begin{itemize}
    \item Stack the green block onto the yellow block.
    \item Place the carrot on the plate.
    \item Put the spoon on the towel.
    \item Move the eggplant from the sink to the basket.
\end{itemize}

During the experiments, we did not perform any prompt fine-tuning for specific tasks; the prompts used were identical to those in the physical experiments. Task instructions were issued in a format consistent with that of SimplerEnv. Additionally, we only utilized RGB data and depth data provided by the simulation platform, without employing any privileged information such as object positions or object boundary information.

In the experiments, we took 24 attempts as a unit, and each task was repeated 3 times with different random seeds. The results of each experiment were obtained by averaging the repeated trials.

\subsection{Results of Simulation}

\begin{table}[htbp]
\vspace{-0em}
\caption{Results of ManiAgent in simulation experiments}
\vspace{-0.5em}
\label{tab:simulation}
\resizebox{\linewidth}{!}{
\begin{tabular}{cccccc}
\hline
Method & Task 1 & Task 2 & Task 3 & Task 4 & Average \\ \hline
CogACT~\cite{li2024cogact} & 15.0\% & 50.8\% & 71.7\% & 67.5\% & 51.3\% \\
\rowcolor[HTML]{EFEFEF} 
pi-0~\cite{black2024pi_0} & 21.3\% & 58.8\% & 63.3\% & \textbf{79.2\%} & 55.7\% \\
ManiAgent-GPT-4o & 76.4\% & {\underline{95.8\%}} & 77.8\% & 47.2\% & 74.3\% \\
\rowcolor[HTML]{EFEFEF} 
ManiAgent-GPT-5-nano & 72.2\% & 63.9\% & 62.5\% & 50.0\% & 62.2\% \\
ManiAgent-GPT-5 & {\underline{87.5\%}} & {\underline{95.8\%}} & \textbf{91.7\%} & {\underline{72.2\%}} & \textbf{86.8\%} \\
\rowcolor[HTML]{EFEFEF} 
ManiAgent-Claude-4-sonnet & 77.8\% & \textbf{98.6\%} & 80.6\% & 62.5\% & 79.9\% \\
ManiAgent-Grok-4 & \textbf{88.9\%} & \textbf{98.6\%} & {\underline{83.3\%}} & 61.1\% & {\underline{83.0\%}} \\ \hline
\end{tabular}
}
\vspace{-0em}
\end{table}

The experimental results are shown in Table~\ref{tab:simulation}. ManiAgent achieves high success rates across all SimplerEnv tasks and, in aggregate, substantially surpasses representative VLA (CogACT~\cite{li2024cogact} and Pi\mbox{-}0~\cite{black2024pi_0}). Moreover, performance scales with the capability of the VLM, indicating that our framework effectively converts stronger perception\mbox{-}language\mbox{-}reasoning capacity into higher action success. With GPT\mbox{-}5, ManiAgent attains an average success rate of 86.8\% over the four tasks, clearly outperforming prior methods. 

At the per\mbox{-}task level, ManiAgent remains ahead on three of the four tasks; the only exception is Task~4 (``move the eggplant from the sink to the basket''), where certain initial placements cause partial occlusion by the sink. In these cases, our detection module can mistakenly localize the sink rim as the eggplant center, leading to grasp failures; this failure mode is addressable via detector\mbox{-}specific adjustments (e.g., class\mbox{-}aware masking or depth\mbox{-}validated center refinement). In addition, intermittent depth\mbox{-}RGB misalignment in the simulator further limits observed performance. Consequently, the reported numbers are conservative and likely underestimate the true capability of our framework.

\subsection{Physical Experimental Setup}

\begin{figure*}[htbp]
    \centering
    \includegraphics[width=0.85\linewidth]{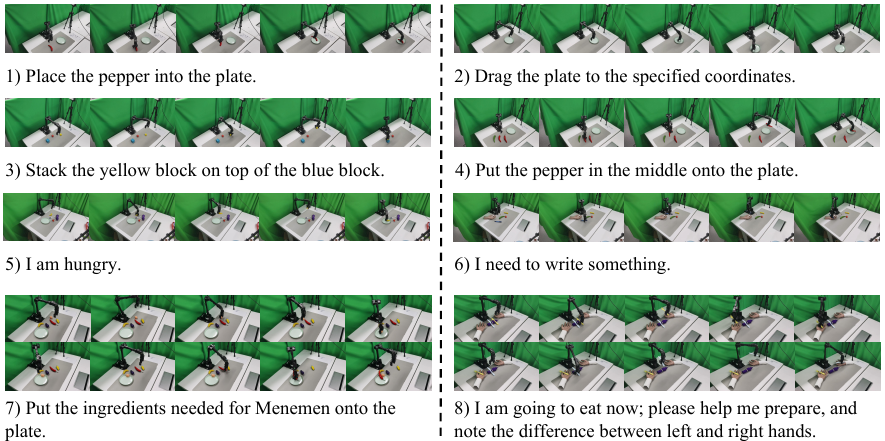}
    \vspace{-0.5em}
    \caption{Definition and scenario examples of real-world manipulation tasks}
    \label{fig:real_overall}
    \vspace{-1em}
\end{figure*}

We tested our method on a physical platform, using the WidowX-250s robotic arm, which is suitable for algorithm validation. In the experimental scene, we employed two Realsense D435 cameras to generate point clouds, and obtained grasping poses through point cloud processing. Among them, the RGB images captured by the main-view camera were used as the input perspective for reasoning by each agent.

We adopted Florence-v2~\cite{xiao2024florence} as the object detection algorithm, and uniformly used GPT-5 as the scene description tool, which can ensure the correctness of the scene description format and effectively improve experimental efficiency. Since we need to test the performance of each model on unseen tasks, the cache function of the models was disabled during this part of the experiment.

To evaluate the capability of our method, we defined the following task evaluation metrics:
\begin{itemize}
    \item \textbf{Non Pick-and-Place Capability(NP)}: Aimed at assessing whether the method can handle tasks beyond simple Pick-and-Place operations.
    \item \textbf{High Generalization Ability(HG)}: Aimed at assessing whether the method has high generalization ability, i.e., can complete new tasks without data fine-tuning for specific tasks.
    \item \textbf{Relative Position Perception Ability(RP)}: Aimed at assessing whether the method can distinguish relative positions, such as ``leftmost'' and ``rightmost''.
    \item \textbf{Intention Reasoning Ability(IR)}: Aimed at assessing whether the method can complete tasks described in an indirect manner, e.g., inferring users' real needs from vague instructions.
    \item \textbf{Knowledge Retrieval and Utilization Ability(KR)}: Aimed at assessing whether the method can retrieve and utilize common-sense information.
    \item \textbf{Multi-step Task Planning Ability(MP)}: Aimed at assessing the method's ability to execute multi-step tasks, such as whether it suffers from step forgetting.
\end{itemize}

To comprehensively evaluate our method's capabilities, we designed 8 tasks with corresponding prompts as Fig.~\ref{fig:real_overall}.

\begin{table}[htbp]
\centering
\caption{Characteristics of physical experimental tasks}
\vspace{-0.5em}
\label{tab:task_define}
\begin{tabular}{cccccll}
\hline
Task                       & NP                   & HG                   & RP                   & IR         & KR                             & MP                             \\ \hline
Task 1                     &                      &                &                      &            &                                &                                \\
Task 2                     & \checkmark           &                      &                      &            & \multicolumn{1}{c}{\checkmark} &                                \\
Task 3                     &                & \checkmark           & \checkmark           & \textbf{}  & \textbf{}                      &                                \\
Task 4                     &                      & \textbf{}            & \checkmark           &      & \multicolumn{1}{c}{\checkmark} &                                \\
\multicolumn{1}{l}{Task 5} & \multicolumn{1}{l}{} & \multicolumn{1}{l}{} & \multicolumn{1}{l}{} & \checkmark &                                &                                \\
\multicolumn{1}{l}{Task 6} & \multicolumn{1}{l}{} & \multicolumn{1}{l}{} & \multicolumn{1}{l}{} & \checkmark &                                &                                \\
\multicolumn{1}{l}{Task 7} & \multicolumn{1}{l}{} & \checkmark           & \multicolumn{1}{l}{} & \checkmark & \multicolumn{1}{c}{\checkmark} & \multicolumn{1}{c}{\checkmark} \\
Task 8                     & \textbf{}            & \textbf{}            & \checkmark           & \checkmark & \multicolumn{1}{c}{\checkmark} & \multicolumn{1}{c}{\checkmark} \\ \hline
\end{tabular}
\vspace{-0.5em}
\end{table}
The correspondence between each task and the aforementioned capability metrics is shown in Table \ref{tab:task_define}.

\subsection{Main Results of Physical Experiment}

\begin{table*}[htbp]
\centering
\caption{The main results of the physical experiment}
\label{tab:main_result_exp}
\begin{tabular}{ccccccccccc}
\hline
\multicolumn{2}{c}{Method} & Task 1 & Task 2 & Task 3 & Task 4 & Task 5 & Task 6 & Task 7 & Task 8 & Average \\ \hline
 & GPT-4o & 5/6 & 5/6 & 3/6 & 4/6 & 4/6 & 3/6 & 4/6 & 2/6 & 62.5\% \\
 & \cellcolor[HTML]{EFEFEF}GPT-5-nano & \cellcolor[HTML]{EFEFEF}5/6 & \cellcolor[HTML]{EFEFEF}6/6 & \cellcolor[HTML]{EFEFEF}4/6 & \cellcolor[HTML]{EFEFEF}6/6 & \cellcolor[HTML]{EFEFEF}4/6 & \cellcolor[HTML]{EFEFEF}4/6 & \cellcolor[HTML]{EFEFEF}4/6 & \cellcolor[HTML]{EFEFEF}3/6 & \cellcolor[HTML]{EFEFEF}75\% \\
 & GPT-5 & 6/6 & 6/6 & 6/6 & 6/6 & 6/6 & 5/6 & 4/6 & 6/6 & 93.8\% \\
 & \cellcolor[HTML]{EFEFEF}Claude-4-sonnet & \cellcolor[HTML]{EFEFEF}6/6 & \cellcolor[HTML]{EFEFEF}6/6 & \cellcolor[HTML]{EFEFEF}6/6 & \cellcolor[HTML]{EFEFEF}5/6 & \cellcolor[HTML]{EFEFEF}6/6 & \cellcolor[HTML]{EFEFEF}6/6 & \cellcolor[HTML]{EFEFEF}6/6 & \cellcolor[HTML]{EFEFEF}5/6 & \cellcolor[HTML]{EFEFEF}95.8\% \\
\multirow{-5}{*}{\begin{tabular}[c]{@{}c@{}}Commercial \\ VLM\end{tabular}} & Grok-4 & 6/6 & 6/6 & 4/6 & 6/6 & 6/6 & 6/6 & 6/6 & 6/6 & 95.8\% \\ \hline
 & \cellcolor[HTML]{EFEFEF}GPT-oss-120b & \cellcolor[HTML]{EFEFEF}4/6 & \cellcolor[HTML]{EFEFEF}6/6 & \cellcolor[HTML]{EFEFEF}2/6 & \cellcolor[HTML]{EFEFEF}4/6 & \cellcolor[HTML]{EFEFEF}4/6 & \cellcolor[HTML]{EFEFEF}2/6 & \cellcolor[HTML]{EFEFEF}2/6 & \cellcolor[HTML]{EFEFEF}1/6 & \cellcolor[HTML]{EFEFEF}52.1\% \\
\multirow{-2}{*}{\begin{tabular}[c]{@{}c@{}}Opensource \\ VLM\end{tabular}} & Qwen-3-235b & 6/6 & 5/6 & 1/6 & 6/6 & 3/6 & 2/6 & 1/6 & 2/6 & 54.2\% \\ \hline
\end{tabular}
\vspace{-0.5em}
\end{table*}

The main experimental results are summarized in Table~\ref{tab:main_result_exp}. Our framework successfully completed all tasks across all VLMs, with Claude-4-sonnet and Grok-4 achieving the highest average success rate of 95.8\%. In contrast, open-source models performed significantly worse. This is mainly due to open-source models struggling with format adherence, where token loss during output generation leads to invalid action formats and task failures, an issue less prevalent in commercial models.

Several key experimental phenomena were observed:

For Task 3, the height of blocks caused difficulty in grasping, as the object detection algorithm often localized center points on the block sides. This issue required using grasp pose coordinates generated by the grasp pose generator. All models except GPT-4o successfully performed using the generated coordinates.

In Task 4, combinations of green and red chili peppers interfered with the object selection function, causing Claude-4-sonnet to confuse the objects to be grasped.

For Task 7, GPT-series VLMs occasionally selected ingredients from the recipe not present in the scene (e.g., tomatoes), resulting in task failure.

In Task 8, GPT-4o placed utensils incorrectly, ignoring the convention of holding a fork in the left hand and a knife in the right, while other models performed the task correctly.

In summary, our method achieved a high success rate in real-world experiments. The performance of the framework is closely related to the capabilities of the underlying VLM, demonstrating that the framework fully leverages the potential of the VLMs used.

\subsection{Performance Comparison}

\begin{table*}[htbp]
\centering
\caption{Comparison of physical scenario performance between ManiAgent and ReKep\cite{huang2024rekep}}
\label{tab:rekep}
\begin{tabular}{cccccccccccc}
\hline
\multicolumn{2}{c}{} & \multicolumn{2}{c}{Task 1} & \multicolumn{2}{c}{Task 3} & \multicolumn{2}{c}{Task 5} & \multicolumn{2}{c}{Task 7} & \multicolumn{2}{c}{Average} \\
\multicolumn{2}{c}{\multirow{-2}{*}{VLM}} & ReKep & \textbf{Ours} & ReKep & \textbf{Ours} & ReKep & \textbf{Ours} & ReKep & \textbf{Ours} & ReKep & \textbf{Ours} \\ \hline
 & GPT-4o & 3/6 & 5/6 & 1/6 & 3/6 & 2/6 & 4/6 & 0/6 & 4/6 & 25\% & 66.7\% \\
 & \cellcolor[HTML]{EFEFEF}GPT-5-nano & \cellcolor[HTML]{EFEFEF}2/6 & \cellcolor[HTML]{EFEFEF}5/6 & \cellcolor[HTML]{EFEFEF}0/6 & \cellcolor[HTML]{EFEFEF}4/6 & \cellcolor[HTML]{EFEFEF}1/6 & \cellcolor[HTML]{EFEFEF}4/6 & \cellcolor[HTML]{EFEFEF}0/6 & \cellcolor[HTML]{EFEFEF}4/6 & \cellcolor[HTML]{EFEFEF}12.5\% & \cellcolor[HTML]{EFEFEF}70.8\% \\
 & GPT-5 & 4/6 & 6/6 & 0/6 & 6/6 & 2/6 & 6/6 & 0/6 & 4/6 & 25\% & 91.7\% \\
 & \cellcolor[HTML]{EFEFEF}Claude-4-sonnet & \cellcolor[HTML]{EFEFEF}2/6 & \cellcolor[HTML]{EFEFEF}6/6 & \cellcolor[HTML]{EFEFEF}1/6 & \cellcolor[HTML]{EFEFEF}6/6 & \cellcolor[HTML]{EFEFEF}3/6 & \cellcolor[HTML]{EFEFEF}6/6 & \cellcolor[HTML]{EFEFEF}0/6 & \cellcolor[HTML]{EFEFEF}6/6 & \cellcolor[HTML]{EFEFEF}25\% & \cellcolor[HTML]{EFEFEF}100\% \\
\multirow{-5}{*}{\begin{tabular}[c]{@{}c@{}}Commercial\\ VLM\end{tabular}} & Grok-4 & 3/6 & 6/6 & 1/6 & 4/6 & 3/6 & 6/6 & 0/6 & 6/6 & 29.2\% & 91.7\% \\ \hline
 & \cellcolor[HTML]{EFEFEF}GPT-oss-120b & \cellcolor[HTML]{EFEFEF}2/6 & \cellcolor[HTML]{EFEFEF}4/6 & \cellcolor[HTML]{EFEFEF}0/6 & \cellcolor[HTML]{EFEFEF}2/6 & \cellcolor[HTML]{EFEFEF}1/6 & \cellcolor[HTML]{EFEFEF}4/6 & \cellcolor[HTML]{EFEFEF}0/6 & \cellcolor[HTML]{EFEFEF}2/6 & \cellcolor[HTML]{EFEFEF}12.5\% & \cellcolor[HTML]{EFEFEF}50\% \\
\multirow{-2}{*}{\begin{tabular}[c]{@{}c@{}}Opensource\\ VLM\end{tabular}} & Qwen-3-235b & 1/6 & 6/6 & 0/6 & 1/6 & 0/6 & 3/6 & 0/6 & 1/6 & 4.2\% & 45.8\% \\ \hline
\end{tabular}
\vspace{-2em}
\end{table*}

For a horizontal comparison, we evaluate our method against ReKep~\cite{huang2024rekep} on Tasks 1, 3, 5, and 7 under the same settings. The results are summarized in Table~\ref{tab:rekep}.

Overall, our framework significantly outperforms ReKep across all four tasks. When using the same VLM, ManiAgent achieves success rate improvements ranging from 37.5\% (GPT-oss-120b) to 75\% (Claude-4-sonnet) compared to ReKep. This demonstrates the effectiveness of our approach in leveraging VLMs for improved task execution.

For Task 1, a simple pick-and-place task, the difference in performance between our method and ReKep is not as pronounced, as both methods can complete the task with high success rates. 

In Task 3, the inability of ReKep to use target grasp poses for trajectory optimization results in poor performance. Specifically, ReKep cannot generate an effective target position directly above the block to be stacked, leading to failure in the stacking task. 

For Task 5, the complexity of the scene, with numerous objects in close proximity, causes ReKep to generate keypoints that are often not aligned with the target object. This misalignment leads to grasp failures, as ReKep struggles to identify the correct grasp pose. In contrast, ManiAgent, with its more advanced reasoning and pose generation, performs significantly better in such complex scenes.

In Task 7, ReKep’s lack of long-horizon reasoning capabilities is a critical limitation. Without the ability to plan and execute complex, multi-step tasks, ReKep fails to complete this task, whereas ManiAgent successfully handles the multi-step reasoning required for the task.

These comparisons highlight the advantages of ManiAgent’s approach, especially in handling more complex tasks that require accurate pose estimation and long-horizon planning, areas where ReKep falls short due to its limitations in task planning and reasoning.

\subsection{Automatic Data Collection}

\begin{figure}[htbp]
    \centering
    \vspace{-0.5em}
    \includegraphics[width=1\linewidth]{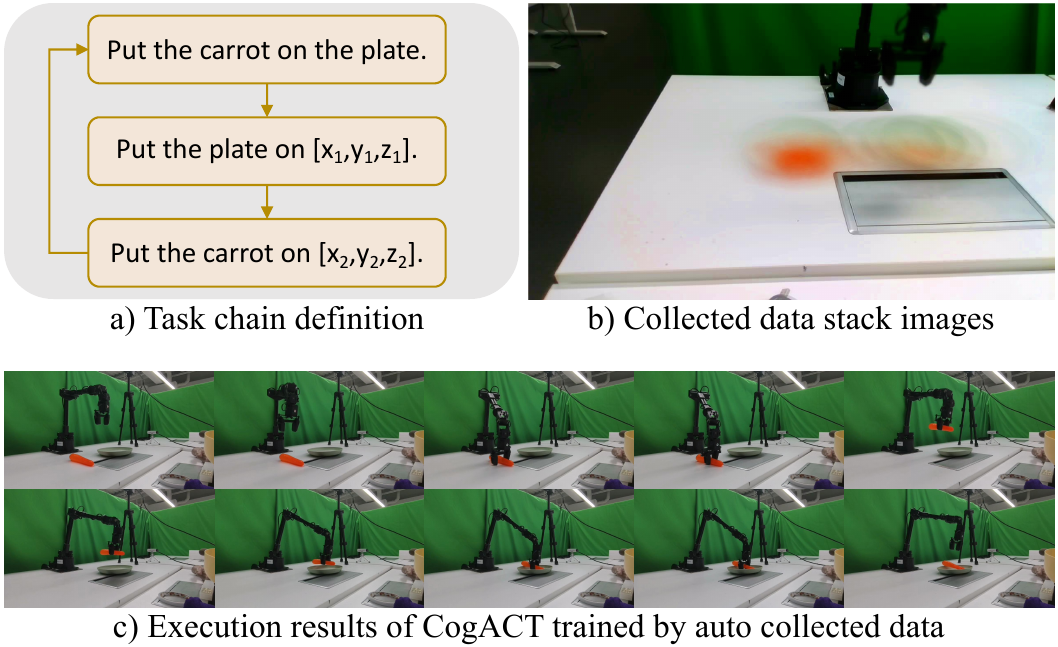}
    \vspace{-2.0em}
    \caption{ManiAgent automates data collection: a) Task chain definition for placing carrots on a plate with coordinates generated using a rule-based method; b) RGB stack of the first frame of collected data, showing objects initially aligned along a line as per the collection strategy; c) Performance of CogACT, trained on data collected by our method, in real-world deployment.}
    \label{fig:auto_data}
    \vspace{-0.5em}
\end{figure}

The scarcity of high-quality training data remains a critical bottleneck for learning-based robotic systems. Recent work has explored automated data collection to scale up robot learning. AutoRT~\cite{ahn2024autort} presents an embodied foundation model for large-scale orchestration of robotic agents, demonstrating how LLMs can coordinate multiple robots for data collection. RoboCat~\cite{bousmalis2024robocat} introduces a self-improving generalist agent that can adapt to new tasks with minimal demonstrations and automatically generate training data through self-supervised learning. These automated collection systems primarily focus on either task diversity or self-improvement, but often require substantial human oversight for scene resetting and failure recovery.

Owing to the high reliability of ManiAgent, our method is well-suited for serving as an automatic data collection approach, enabling end-to-end automated data acquisition from task environment setup to data collection execution.

The data collection process aligns with the framework described earlier, where the key to efficient automatic collection lies in reliable scene reset. During this process, reset can be implemented via either random or rule-based methods. Meanwhile, we have integrated the dataset recording functionality into the controller.

Notably, since our method allows defining tasks using coordinates, we can generate coordinates of objects to be collected in the next step through rule-based approaches. This capability enables our method to uniformly collect operation data for every position in the scene, thereby enhancing the positional diversity of the dataset.

Given the long latency associated with VLM calls, we adopted a rule-based approach to determine task success: a task is deemed successfully executed if the distance between the object's final position and the target position is less than 15 cm; otherwise, the task is re-executed.

In practical experiments, we performed data collection for the task \textit{Place the carrot on the plate}, utilizing parameterized action cache sequences generated by GPT-5. A total of 551 trajectories were collected, among which 450 were valid, resulting in a task success rate of 81.51\%. The total time consumed was 19.5 hours, with an average of 2 minutes per trajectory. Manual intervention was required 15 times during the collection process(an average of one intervention every 46 minutes), primarily due to unexpected movements of the robotic arm-caused by the failure of inverse kinematics to plan a reasonable path, which moved the target object out of the operable range. In such cases, manual scene restoration was necessary.

We attempted to train and deploy a model using the collected dataset with CogACT \cite{li2024cogact}. As shown in Figure \ref{fig:auto_data}, our dataset can train a VLA capable of normal action execution, which validates the effectiveness of our method for fully automatic data collection.

%% file: sections/5_con.tex
\section{Conclusions}

In this paper, we introduce ManiAgent, a framework that decomposes general manipulation tasks into four stages, using specialized agents for perception, reasoning, and control to complete robotic operations. Our experimental results demonstrate that ManiAgent outperforms most VLA models in simulation, achieving an 86.8\% success rate, and attains an average success rate of 95.8\% in real-world tasks when equipped with a high-capability VLM. Additionally, ManiAgent’s high success rate in general manipulation makes it an effective tool for automated data collection, enabling the generation of high-quality datasets for VLA training at reduced costs. Future work will focus on enhancing real-time feedback, extending applications to various platforms beyond robotic arms, and exploring human-robot interaction.